\documentclass[runningheads,a4paper]{llncs}

\usepackage{times}

\usepackage{amssymb}
\usepackage{amsmath}
\usepackage{graphicx}

\usepackage[sf,SF]{subfigure}

\usepackage[numbers]{natbib}

\usepackage{multirow}
\usepackage{xcolor,colortbl}

\usepackage{url}
\urldef{\mailsa}\path|{alfred.hofmann, ursula.barth, ingrid.haas, frank.holzwarth,|
\urldef{\mailsb}\path|anna.kramer, leonie.kunz, christine.reiss, nicole.sator,|
\urldef{\mailsc}\path|erika.siebert-cole, peter.strasser, lncs}@springer.com|    
\newcommand{\keywords}[1]{\par\addvspace\baselineskip
\noindent\keywordname\enspace\ignorespaces#1}

\newcommand{\mysection}[1]{\section{\uppercase{\sffamily#1}}}
\newcommand{\mysubsection}[1]{\subsection{\uppercase{\sffamily#1}}}
\newcommand{\mysubsubsection}[1]{\textbf{\sffamily \uppercase{#1}}\quad  }

\newcommand{\myparagraph}[1]{\textbf{\sffamily #1}\quad  }

\usepackage[pdftitle={Deep Semantic (KR 2016 -- Field Reports)},
pdfauthor={Jakob Suchan, Mehul Bhatt},
linkbordercolor=white,
bookmarksopen=false,
bookmarksnumbered=false]{hyperref}

\hypersetup{
    pdftitle={ILP 2016}, % title
    pdfauthor={Jakob Suchan, Mehul Bhatt, Carl Schultz},     	% author
    pdfsubject={TODO}, % subject of the document
    pdfkeywords={TODO} {TODO} {TODO}, 					% list of keywords	eator of the document
    unicode=false,          									% non-Latin characters in AcrobatÕs bookmarks
    pdftoolbar=false,        									% show AcrobatÕs toolbar?
    pdfmenubar=true,        								% show AcrobatÕs menu?
    pdffitwindow=false,     									% window fit to page when opened
    pdfstartview={FitH},    								% fits the width of the page to the window
    pdfnewwindow=true,      								% links in new window
    bookmarks=true,         								% show bookmarks bar?    
    colorlinks=true,       									% false: boxed links; true: colored links
    linkcolor=red!70!black,          							% color of internal links (change box color with linkbordercolor)
    citecolor=blue!70!black,        							% color of links to bibliography
    filecolor=blue,      									% color of file links
    urlcolor=blue!70!black          								% color of external links
}

\begin{document}

\mainmatter  % start of an individual contribution

\title{Deeply Semantic Inductive Spatio-Temporal Learning}
%\title{Deeply Semantic Spatio-Temporal Learning}

%\title{`Deeply' Semantic Spatio-Temporal Learning}
% first the title is needed
%\title{Inductively Learning\\`Deeply' Semantic Spatio-Temporal Relations}
%\title{ILP(QS) -- A General Inductive Spatial Learning Framework}

% a short form should be given in case it is too long for the running head
\titlerunning{Deeply Semantic Spatio-Temporal Learning}
%\titlerunning{Inductively Learning `Deeply' Semantic Spatio-Temporal Relations}

% the name(s) of the author(s) follow(s) next
%
% NB: Chinese authors should write their first names(s) in front of
% their surnames. This ensures that the names appear correctly in
% the running heads and the author index.
%

%

% (feature abused for this document to repeat the title also on left hand pages)

\author{\small\sffamily Jakob Suchan$^1$ \and Mehul Bhatt$^1$ \and Carl Schultz$^2$}

\authorrunning{J. Suchan, M. Bhatt, C. Schultz}

%\author{Jakob Suchan$^1$ \and Mehul Bhatt$^1$ \and Carl Schultz$^2$}

% the affiliations are given next; don't give your e-mail address
% unless you accept that it will be published

%\institute{\small\sffamily Human-Centred Cognitive Assistance$^1$ \href{http://hcc.uni-bremen.de}{hcc.uni-bremen.de}\sffamily\ and\\\small\sffamily The DesignSpace Group.,$^{1,2}$ \href{http://www.design-space.org}{www.design-space.org}\small\sffamily\  \\University of Bremen$^1$, and University of M\"{u}nster$^2$, GERMANY}

\institute{\small\sffamily Human-Centred Cognitive Assistance$^1$ \href{http://hcc.uni-bremen.de}{hcc.uni-bremen.de}\sffamily\ and\\\small\sffamily The DesignSpace Group.,$^{1,2}$ \href{http://www.design-space.org}{www.design-space.org}\small\sffamily\  \\University of Bremen$^1$, and University of M\"{u}nster$^2$, GERMANY\\[8pt]Spatial Reasoning., \href{http://www.spatial-reasoning.com/}{www.spatial-reasoning.com}
}

%\author{Alfred Hofmann%
%\thanks{Please note that the LNCS Editorial assumes that all authors have used
%the western naming convention, with given names preceding surnames. This determines
%the structure of the names in the running heads and the author index.}%
%\and Ursula Barth\and Ingrid Haas\and Frank Holzwarth\and\\
%Anna Kramer\and Leonie Kunz\and Christine Rei\ss\and\\
%Nicole Sator\and Erika Siebert-Cole\and Peter Stra\ss er}
%
%\authorrunning{Bhatt, Suchan, Schultz}
% (feature abused for this document to repeat the title also on left hand pages)

% the affiliations are given next; don't give your e-mail address
% unless you accept that it will be published
%\institute{Springer-Verlag, Computer Science Editorial,\\
%Tiergartenstr. 17, 69121 Heidelberg, Germany\\
%\mailsa\\
%\mailsb\\
%\mailsc\\
%\url{http://www.springer.com/lncs}}

%
% NB: a more complex sample for affiliations and the mapping to the
% corresponding authors can be found in the file "llncs.dem"
% (search for the string "\mainmatter" where a contribution starts).
% "llncs.dem" accompanies the document class "llncs.cls".
%

\toctitle{Deeply Semantic Inductive Spatio-Temporal Learning}
\tocauthor{J. Suchan, M. Bhatt, C. Schultz}
\maketitle

%a mixed qualitative-quantitative
%We present a mixed qualitative-quantitative spatio-temporal learning framework rooted in inductive logic programming. The framework and  resulting system support learning with relational spatio-temporal features identifiable in a range of domains involving the processing and interpretation of dynamic visuo-spatial imagery at the interface of visuo-spatial language, logic, and cognition. We report on the prototypical system, and present an example involving its use in the domain of computing for the domain of visual arts, social sciences, and computational cognitive science.

\begin{abstract}
We present an inductive spatio-temporal learning framework rooted in inductive logic programming. With an emphasis on visuo-spatial language, logic, and cognition, the framework supports learning with relational spatio-temporal features identifiable in a range of domains involving the processing and interpretation of dynamic visuo-spatial imagery.  We present a prototypical system, and an example application in the domain of computing for visual arts and computational cognitive science.

\keywords{Spatio-Temporal Learning; Dynamic Visuo-Spatial Imagery; Declarative Spatial Reasoning; Inductive Logic Programming; AI and Art}
\end{abstract}

\mysection{Introduction}
Cognitive assistive technologies and computer-human interaction systems involving an interplay of space, dynamics, and cognition necessitate capabilities for explainable reasoning, learning, and control  about \emph{space, actions, change}, and \emph{interaction} \citep{Bhatt:RSAC:2012}. Prime application scenarios, for instance, include {\footnotesize(A1--A5)}:\quad  {\footnotesize\bf (A1).} activity grounding from video and point-clouds; {\footnotesize\bf (A2).} modelling and analysis of environmental processes at the geospatial scale; {\footnotesize\bf (A3).} medical computing scenarios replete with visuo-spatial imagery; {\footnotesize\bf (A4).} visuo-locomotive human behavioural data concerning aspects such as mobility or navigation, eye-tracking based visual perception research; {\footnotesize\bf (A5).} embodied human-machine interaction and control for commonsense cognitive robotics. A crucial requirement in relevant application contexts (such as {\small A1--A5}) pertains to the semantic interpretation of multi-modal human behavioural or socio-environmental data, with objectives ranging from knowledge acquisition (e.g., medical computing, computer-aided learning) and data analyses (e.g., activity intepretation) to hypothesis formation in experimental settings (e.g., empirical visual perception studies). The focus of our research is the processing and interpretation of dynamic visuo-spatial imagery with a particular emphasis on the ability to learn commonsense knowledge that is semantically founded in spatial, temporal, and spatio-temporal relations and patterns.

%
%
%\smallskip
%
%\begin{itemize}{\small
%	\item [{\footnotesize (A1).}] activity grounding from video and point-clouds
%	\item [{\footnotesize (A2).}] modelling and analysis of environmental processes at the geospatial scale
%	\item [{\footnotesize (A3).}] medical computing scenarios replete with visuo-spatial imagery
%	\item [{\footnotesize (A4).}] visuo-locomotive human behavioural data concerning aspects such as mobility or\\navigation, eye-tracking based visual perception research
%	\item [{\footnotesize (A5).}] embodied human-machine interaction and control for commonsense cognitive robotics
%	}
%\end{itemize}

\medskip

\mysubsubsection{Deep Visuo-Spatial Semantics}\quad  The high-level semantic interpretation and qualitative analysis of dynamic visuo-spatial imagery requires the representational and inferential mediation of commonsense abstractions of \emph{space, time, action, change, interaction} and their mutual interplay thereof. In this backdrop, \emph{deep visuo-spatial semantics} denotes the existence of declaratively grounded models ---e.g., pertaining to \emph{space, time, space-time, motion, actions \& events, spatio-linguistic conceptual knowledge}--- and systematic formalisation supporting capabilities such as:\quad \textbf{\small(a)}. mixed quantitative qualitative spatial inference and question answering (e.g., about consistency, qualification and quantification of relational knowledge);\quad \textbf{\small(b)}. non-monotonic spatial reasoning	(e.g., for abductive explanation);\quad 	\textbf{\small(c)}. relational learning of spatio-temporally grounded concepts;\quad \textbf{\small(d)}. integrated inductive-abductive spatio-temporal inference;\quad \textbf{\small(e)}. probabilistic spatio-temporal inference;\quad \textbf{\small(f)}. embodied grounding and simulation from the viewpoint of cognitive linguistics (e.g., for knowledge acquisition and inference based on natural language).

%\begin{itemize}
%	\item mixed quantitative-qualitative spatial inference and question answering (e.g., about consistency, qualification and quantification of relational knowledge)
%	\item non-monotonic spatial reasoning	(e.g., for abductive explanation)	
%	\item relational learning of spatio-temporally grounded concepts
%	\item integrated inductive-abductive spatio-temporal inference
%	\item probabilistic spatio-temporal inference
%	\item embodied grounding and simulation from the viewpoint of cognitive linguistics (e.g., for knowledge acquisition and inference based on natural language)
%
%\end{itemize}

%DBLP:conf/kr/HaarslevLM98

Recent perspectives on deep visuo-spatial semantics encompass methods for declarative (spatial) representation and reasoning ---e.g., about \emph{space and motion}--- within frameworks such as constraint logic programming (rule-based spatio-temporal inference \citep{clpqs-cosit11,eccv2014}), answer-set programming (for non-monotonic spatial reasoning \citep{ampmtqs-lpnmr2015}), description logics (for spatio-terminological reasoning \citep{cosit09/BhattDH09}), inductive logic programming (for inductive-abductive spatio-temporal learning \citep{ilp/DubbaBDHC12,jair/DubbaCHBD15}) and other specialised forms of commonsense reasoning based on expressive action description languages for modelling \emph{space, events, action, and change} \citep{Bhatt:RSAC:2012,bhatt:scc:08}. In general, deep visuo-spatial semantics driven by declarative spatial representation and reasoning pertaining to dynamic visuo-spatial imagery is relevant and applicable in a variety of cognitive interaction systems and assistive technologies at the interface of (spatial) language, (spatial) logic, and (visuo-spatial) cognition.

%encompassing other areas such as geographic information systems, cognitive robotics, computer-aided education / learning.

\medskip

%\mysubsubsection{Space and Time:\quad  Inductive Learning with Deep Semantics}\quad 
\mysubsubsection{Inductive Spatio-Temporal Learning (with Deep Semantics)} \\
This research is motivated by the need to have a systematic inductive logic programming \citep{Muggleton94inductivelogic} founded spatio-temporal learning framework and corresponding system that:

%{\footnotesize ---}\quad provides an expressive spatio-linguistically motivated ontology to predicate primitive and complex (domain-independent) relational spatio-temporal features identifiable in a broad range of application domains (e.g., {\small A1--A5}) involving the processing and interpretation of dynamic visuo-spatial imagery.
%
%	
%{\footnotesize ---}\quad supports spatio-temporal relations natively such that the semantics of these relations is directly built into the underlying ILP-based learning framework.
%
%
%{\footnotesize ---}\quad supports seamless mixing of, and transition between, quantitative and qualitative spatial data.
\begin{itemize}
		
\item[{\footnotesize --}] provides an expressive spatio-linguistically motivated ontology to predicate primitive and complex (domain-independent) relational spatio-temporal features identifiable in a broad range of application domains (e.g., {\small A1--A5}) involving the processing and interpretation of dynamic visuo-spatial imagery.

\smallskip
	
\item[{\footnotesize --}] supports spatio-temporal relations natively such that the semantics of these relations is directly built into the underlying ILP-based learning framework.

\smallskip

\item[{\footnotesize --}] supports seamless mixing of, and transition between, quantitative and qualitative spatial data.
\end{itemize}

%\begin{itemize}
%		\item  [{\footnotesize (M1).}] provides an expressive spatio-linguistically motivated ontology to predicate primitive and complex (domain-independent) relational spatio-temporal features identifiable in a broad range of application domains (e.g., {\small A1--A5}) involving the processing and interpretation of dynamic visuo-spatial imagery
%	
%	\item  [{\footnotesize (M2).}] supports spatio-temporal relations natively such that the semantics of these relations is directly built into the underlying ILP-based learning framework
%
%	\item  [{\footnotesize (M3).}] supports seamless mixing of, and transition between, quantitative and qualitative spatial data
%
%\end{itemize}

We particularly emphasise and ensure compatibility with the general setup of (constraint) logic programming framework such that diverse knowledge sources and reasoning mechanisms outside of inductive learning may be directly interfaced, and reasoning / learning capabilities be combined within large-scale integrated systems for cognitive computing.

\smallskip

%\textbf{\sffamily System Implementation and Application.}\quad We present a working prototype for an inductive spatio-temporal learning system founded on the Aleph ILP system \citep{aleph-ilp-system}. As a working example, we demonstrate the spatio-temporal learning capability of our framework from the viewpoint of AI-based computing for the arts and social sciences, and computational cognitive science. Aimed at cognitive film studies and visual perception research, the particular use-case presented pertains to the (visual) learning of cinematographic patterns of symmetry and its visual reception (by means of eye-tracking) by subjects. The presented example translates to a variety of cases involving visual perception and human behaviour studies.

\mysection{Learning From Relational Spatio-Temporal Structure: A General Framework and System}
%\mysection{ILP(QS) -- A General Framework for Learning Relational Spatio-Temporal Structure}
%\mysection{ILP(QS) -- A General Framework for Learning Relational Spatio-Temporal Structure}
{We present a general framework and working prototype for an inductive spatio-temporal learning system with an elaborate ontology supporting a range of space-time features; we demonstrate the functional capabilities from the viewpoint of AI-based computing for the arts \& social sciences, and computational cognitive science.}

\mysubsection{The Spatio-Temporal Domain $\mathcal{O}_{sp}$, and $\mathcal{QS}$}
The spatio-temporal ontology $\mathcal{O}_{sp}$$~\equiv_{def}~<$$\mathcal{E}, \mathcal{R}>$ is characterised by the basic spatial entities ($\mathcal{E}$) that can be used as abstract representations of domain-objects and the relational spatio-temporal structure ($\mathcal{R}$) that characterises the qualitative spatio-temporal relationships amongst the supported entities in ($\mathcal{E}$). The following primitive spatial entities are sufficient to characterise the learning mechanism and its sample application for this paper:\quad 

%\mysubsubsection{Spatial Entities ($\mathcal{E}$)}\quad  The following primitive spatial entities are sufficient to characterise the learning mechanism and its sample application for this paper:

\smallskip

{\small a \textbf{point} is a pair of reals $x,y$; \quad a \textbf{vector} is a pair of reals $v_x,v_y$; \quad an \textbf{oriented point} consists of a point $p$ and a vector $v$; \quad a \textbf{line segment} is a pair of end points $p_1, p_2$ ($p_1 \neq p_2$); \quad a \textbf{rectangle} is a point $p$ representing the bottom left corner, a direction vector $v$ defining the orientation of the base of the rectangle, and a real width and height $w,h$ ($0 < w, 0 < h$); \quad an \textbf{axis-aligned rectangle} is a rectangle with fixed direction vector $v=(1,0)$; \quad a \textbf{circle} is a centre point $p$ and a real radius $r$ ($0 < r$); \quad a \textbf{simple polygon} is defined by a list of $n$ vertices (points) $p_1, \dots, p_n$ (spatially ordered counter-clockwise) such that the boundary is non-self-intersecting, i.e., there does not exist a polygon boundary edge between vertices $p_i, p_{i+1}$ that intersects some other edge $p_j, p_{j+1}$ for all $1 \leq i < j < n$ and $i + 1 < j$.
}

\smallskip

%\begin{itemize}{\small
%	\item a \emph{point} is a pair of reals $x,y$,
%	\item a \emph{vector} is a pair of reals $v_x,v_y$,
%	\item an \emph{oriented point} consists of a point $p$ and a vector $v$,
%	\item a \emph{line segment} is a pair of end points $p_1, p_2$ ($p_1 \neq p_2$),
%	\item a \emph{rectangle} is a point $p$ representing the bottom left corner, a direction vector $v$ defining the orientation of the base of the rectangle, and a real width and height $w,h$ ($0 < w, 0 < h$)	
%	\item an \emph{axis-aligned rectangle} is a rectangle with fixed direction vector $v=(1,0)$,
%	\item a \emph{circle} is a centre point $p$ and a real radius $r$ ($0 < r$),
%	\item a \emph{simple polygon} is defined by a list of $n$ vertices (points) $p_1, \dots, p_n$ (spatially ordered counter-clockwise) such that the boundary is non-self-intersecting, i.e., there does not exist a polygon boundary edge between vertices $p_i, p_{i+1}$ that intersects some other edge $p_j, p_{j+1}$ for all $1 \leq i < j < n$ and $i + 1 < j$.
%	}
%\end{itemize}

Spatio-temporal relationships ($\mathcal{R}$) between the basic entities in $\mathcal{E}$ may be characterised with respect to arbitrary spatial and spatio-temporal domains such as \emph{mereotopology, orientation, distance, size, motion}; Table~\ref{tbl:relations} lists the relevant supported relations from the viewpoint of established spatial abstraction calculi such as the Region Connection Calculus \citep{randell1992spatial}, Rectangle Algebra and Block Algebra \citep{guesgen1989spatial}, LR Calculus \citep{Scivos2004}, Oriented-Point Relation Algebra (OPRA) \citep{moratz06_opra-ecai}, and Space-Time Histories \cite{Hayes:Naive-I, hazarika2005qualitative}.

\medskip

%\enlargethispage{0.5in}
\mysubsubsection{$\mathcal{QS}$ -- Analytic Semantics for $\mathcal{O}_{sp}$}
%\mysubsubsection{Analytic Semantics for $\mathcal{QS}$}
We adopt an analytic approach to spatial reasoning, where the semantics of spatial relations are encoded as polynomial constraints within a (constraint) logic programming setup. The analytic method supports the integration of qualitative and quantitative spatial information, and provides a means for sound, complete and approximate spatial reasoning \cite{clpqs-cosit11}. For example, let axis-aligned rectangles $a,b$ each be defined by a bottom-left vertex $(x_i,y_i)$ and a width and height $w_i, h_i$, for $i \in \{a,b\}$ such that $x_i, y_i, w_i, h_i$ are reals. The relation that $a$ is a \emph{non-tangential proper part} of $b$ corresponds to the polynomial constraint:
{\small$$(x_b < x_a) \wedge (x_a + w_a < x_b + w_b) \wedge (y_b < y_a) \wedge (y_a + h_a < y_b + h_b)$$}
Continuing with the example, this is generalised to arbitrarily oriented rectangles. Determining whether a point is inside an arbitrary rectangle is based on vector projection. Point $p$ is projected onto vector $v$ by taking the dot product: $$(x_p, y_p) \cdot (x_v, y_v) = x_p x_v + y_p y_v.$$
With this approach, the task of determining whether a set of spatial relations is consistent then becomes the task of determining whether a system of polynomial constraints is satisfiable. We emphasise that our approach and framework are not limited to the above entities; a wider class of 2D and 3D spatial entities are supported and may be defined as per domain-specific and computational needs \cite{clpqs-cosit11,SchultzRuleML2016,ampmtqs-lpnmr2015,sum2016}.

\begin{table}[t]
\centering
\scriptsize
%\begin{tabular}{||l|r|r|r|r|r||}
%\begin{tabular}{|l|r|r|r||}
\begin{tabular}{|l|p{15.2 ex}|p{45.2 ex}|p{22 ex}|}
\hline
\textbf{\color{blue!70!black}\textsc{Spatial Domain} ($\mathcal{QS}$)} & \emph{Formalisms}  & \emph{Spatial Relations ({\color{blue!70!black}$\mathcal{R}$})} & \emph{Entities ({\color{blue!90!black}$\mathcal{E}$})} \\
\hline
\hline
%% geometry of solids
\multirow{2}{*}{Mereotopology} & {RCC-5, RCC-8 \citep{randell1992spatial}} & {\tiny\sffamily disconnected (dc), external contact (ec), partial overlap (po), tangential proper part (tpp), non-tangential proper part (ntpp), proper part (pp), part of (p), discrete (dr), overlap (o), contact (c)} & arbitrary rectangles, circles, polygons, cuboids, spheres \\
\cline{2-4}
%\multirow{2}{*}{mereotopology} & RCC, RCC-5, RCC-8 & dc, ec, po, tpp, ntpp, eq, pp, p, dr, o, c & rectangles, circles, polygons, cuboids, spheres \\
%\cline{2-4}
&  Rectangle \& Block algebra \citep{guesgen1989spatial} & {\tiny\sffamily proceeds, meets, overlaps, starts, during, finishes, equals} & axis-aligned rectangles and cuboids \\
 \hline
\multirow{2}{*}{Orientation}  & LR \citep{Scivos2004} & {\tiny\sffamily left, right, collinear, front, back, on} & 2D point, circle, polygon with 2D line\\
\cline{2-4}
& OPRA \citep{moratz06_opra-ecai} & {\tiny\sffamily facing towards, facing away, same direction, opposite direction} & oriented points, 2D/3D vectors \\
 \hline
%% square fitting
\multirow{2}{*}{Distance, Size}   & QDC \citep{hernandez1995qualitative} & {\tiny\sffamily adjacent, near, far, smaller, equi-sized, larger} & rectangles, circles, polygons, cuboids, spheres\\
 \hline
\multirow{2}{*}{Dynamics, Motion}   & Space-Time Histories \cite{Hayes:Naive-I, hazarika2005qualitative} & {\tiny\sffamily moving: towards, away, parallel; growing / shrinking:  vertically, horizontally; passing: in front, behind; splitting / merging} & rectangles, circles, polygons, cuboids, spheres\\
\hline
\end{tabular}
% \caption{{\footnotesize\sffamily The Spatio-Temporal Domain $\mathcal{QS}$ Supported within the Learning Framework}}
  \caption{{\footnotesize\sffamily The Spatio-Temporal Domain $\mathcal{O}_{sp}$ supported within the Learning Framework}}
\label{tbl:relations}
\end{table}

\medskip

\mysubsubsection{Inductive Learning with the Spatial System  $<\mathcal{O}_{sp}$,~$\mathcal{QS}>$}
Learning is founded on the Aleph ILP system \citep{aleph-ilp-system}. Learning spatio-temporal structures, is based on integrating the spatial ontology $\mathcal{O}_{sp}$ described above, into the basic learning setup of ILP. 

\smallskip

\textbf{Given:}\quad (1) A set of examples $E$, consisting of positive and negative examples for the desired spatio-temporal structure, i.e., $E = E^+ \cup E^-$, where each example is given by a set of spatio-temporal observations in the domain;
%as a set of basic spatial entities ($\mathcal{E}$) representing domain objects, resp. the relations holding between spatial entities; 
(2) the (spatio-temporal) background knowledge $B$.
%, and (3) the domain independent spatial ontology as defined in $\mathcal{QS}$. 

\smallskip

The \emph{spatio-temporal learning domain} is defined by basic spatial entities ($\mathcal{E}$) constituting the domain objects, the relational spatial structure ($\mathcal{R}$) describing the spatio-temporal configuration of spatial entities in the domain, and rules defining spatio-temporal phenomena and characteristics of the domain.
In this context, spatio-temporal facts characterising the learning examples $E$ can be given as, (a) numerical representation of domain objects, (b) qualitative relations between spatial entities, or (c) a mixed qualitative-quantitative representations,  where the facts are partially grounded in numerical observations. 

%Spatio-temporal phenomena and characteristics of the domain are defined as rules within the background knowledge $B$.
%Further, the background knowledge $B$ defines spatio-temporal rules describing spatio-temporal phenomena in the learning domain.

\medskip

\textbf{Learning:}\quad The learning task is defined as finding hypothesis $H$ consisting of spatio-temporal relations ($\mathcal{R}$) holding between basic spatial entities ($\mathcal{E}$), such that $H \cup B \vDash E^+$, and $H \cup B \nvDash E^-$.

%In this context, spatio-temporal facts can be given as, (a) numerical representation of domain objects, (b) qualitative relations between spatial entities, or (c) a mixed qualitative-quantitative representations,  where the facts are partially grounded in numerical observations. 
As such, the spatial ontology $\mathcal{O}_{sp}$ constitutes an integrated part of the learning setup and spatio-temporal semantics are available throughout the learning process.

%basic spatial entities ($\mathcal{E}$)

%relational spatial structure ($\mathcal{R}$)

%such that $H \cup B \cup \mathcal{QS} \vDash E^+$, and $H \cup B  \cup \mathcal{QS} \nvDash E^-$.

%\textbf{Deep Spatio-Temporal Semantics for Learning} \quad 

%As such, deep spatio-temporal semantics become directly available within the learning process.

% where each example is given as a set of numerical facts describing the geometric arrangement of domain objects

%The spatial domain $\mathcal{QS}$ is integrated into the learning framework by defining a 

%Spatial Domain $\mathcal{QS}$

%basic spatial entities ($\mathcal{E}$)

%relational spatial structure ($\mathcal{R}$)

%\begin{itemize}
%	\item \emph{Spatio-temporal semantics}
%\end{itemize}

%Spatial Domain

%Spatial Primitives

%general spatio-temporal theory

%\begin{itemize}
%	\item \emph{spatio-temporal theory} {$\bf\Sigma_{space}$}
%	\item Domain dependent \emph{spatio-temporal ontology of objects} 
%	\item Domain facts
%\end{itemize}

\smallskip

\mysection{Learning Cinematographic Patterns and their Visual Reception: The Case of  Symmetry}
Aimed at cognitive film studies and visual perception research, we present a use-case pertaining to the (visual) learning of cinematographic patterns of symmetry and its visual reception (by means of eye-tracking) by subjects.\footnote{Our case-study is motivated by a broader multi-level interpretation of symmetry from the viewpoint of film cinematography \citep{Suchan-ACM-SAP-2016}; however, the specific example of this paper focusses on one aspect of this multi-level symmetry characterisation involving relative object placement in a movie scene.} To demonstrate the temporal aspect of the learning framework, we demonstrate the capability to learn ``\emph{axioms of visual perception}'' from dynamic eye-tracking data; both the chosen films and their corresponding eye-tracking data are obtained from a large-scale experiment in visual perception of films \cite{Suchan-Bhatt-WACV2016,Suchan-2016-IJCAI-Visual}. The presented example translates to a variety of cases involving visual perception and human behaviour studies.

%The background knowledge $B$ ... $B = \Sigma_{space} \cup \Sigma_{???} \cup \Sigma_{??}$

\begin{figure}[t]
	\includegraphics[width = \columnwidth]{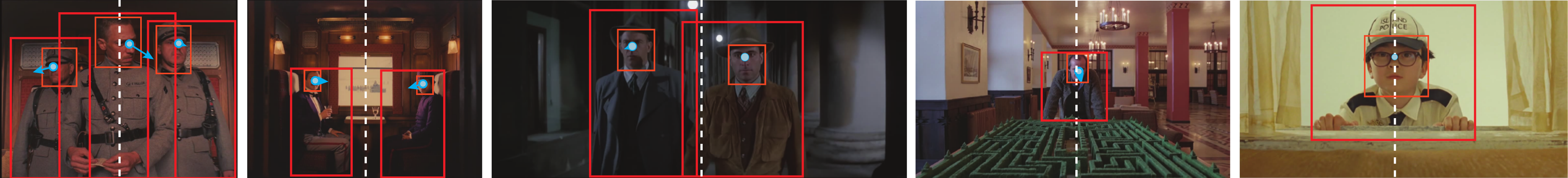}
 \caption{{\footnotesize\sffamily Positive examples for symmetric scene structures at the object level}}
\label{fig:sym-examples}
\end{figure}

%\mysection{Case Study: Cinematographic Patterns and their Visual Reception}

%\begin{itemize}
	%\item 
%\end{itemize}
%
%As an application for learning spatio-temporal rules we examine the field of visuo-spatial perception focussed cognitive film studies, where the key emphasis is on the systematic study and generation of evidence that can characterise and establish correlations between principles for the \emph{synthesis of the moving image}, and its cognitive (e.g., embodied visuo-auditory, emotional) recipient effects on observers \cite{Suchan-2016-WACV-Geometry,Suchan-2016-IJCAI-Visual}.
%
%Symmetry is defined based on a multi-level model \cite{TODO} including the low-level features in the image, the %composition of the frame, and the editing.
%In the context of this paper we focus on symmetry in the configuration of people and objects in the frame.
%

\myparagraph{Learning Spatial Structures: Object-Level Symmetry}
%Towards this we use computer vision algorithms for detection and tracking of people, faces, and their facing directions \cite{Suchan-2016-WACV-Geometry} and extract spatio-temporal primitives representing the geometric configuration of these entities, i.e. in this example we consider \emph{bounding boxes}, \emph{oriented points}, and \emph{line segments}. In this context we focus on the relational spatial configuration of people, and the way in which they are determining the symmetry of the image.
%
As an example for learning spatial structures, we consider symmetry in the relative object placement in a movie scene (see Fig. \ref{fig:sym-examples}). 
In particular, learning is based on the spatial configuration of \emph{people}, \emph{faces}, and their \emph{facing direction}, directly obtained from computer vision algorithms as described in \cite{Suchan-Bhatt-WACV2016}.
In this context, \emph{positive and negative examples}, are given as numerical spatial facts about domain objects in the image.

%\smallskip

%\emph{Data -- Positive and Negative Examples: } 

%As an application for learning spatio-temporal rules we examine the field of visuo-spatial perception focussed cognitive film studies, where the key emphasis is on the systematic study and generation of evidence that can characterise and establish correlations between principles for the \emph{synthesis of the moving image}, and its cognitive (e.g., embodied visuo-auditory, emotional) recipient effects on observers 

\includegraphics[width = 0.99\columnwidth]{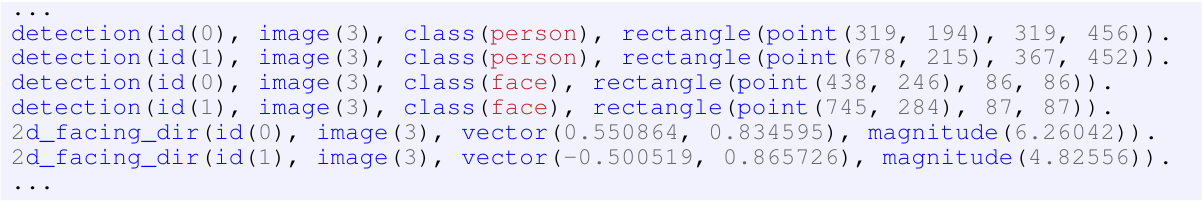}

%\emph{Background Knowledge: } 
We define representations of domain objects linking the numerical description of objects in the image to basic spatial entities describing different aspects of these objects, e.g. the bounding box (\emph{rectangles}), or the center-point (\emph{points}).

\includegraphics[width = \columnwidth]{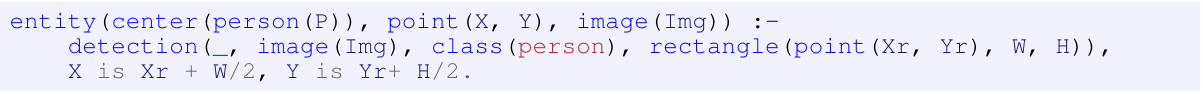}

In addition to the detected domain objects, we define abstract geometric objects needed to describe symmetry, e.g. the \emph{symmetry axis} in the center of the image.

\includegraphics[width = \columnwidth]{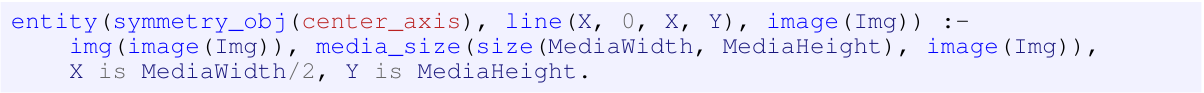}

\emph{Learning: } \quad  
We learn the relational spatial structure consisting of qualitative spatial relationships characterising symmetry in the configuration of the spatial entities in the image, i.e. we consider relations of \emph{topology}, \emph{orientation}, \emph{distance}, and \emph{size}.   

%The spatial structure underlying symmetric images is learned by combining qualitative spatial relations on spatial objects representing the entities of the image, i.e. we include relations on \emph{topology}, \emph{orientation}, \emph{distance}, and \emph{size}.   

\includegraphics[width = \columnwidth]{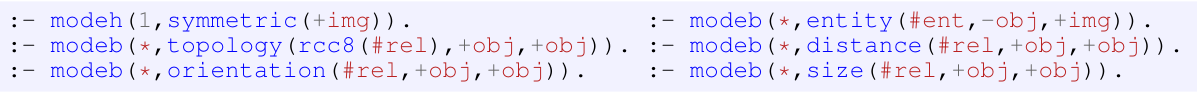}

Exemplary symmetrical spatial structures, learned by the system include the following.

\includegraphics[width = \columnwidth]{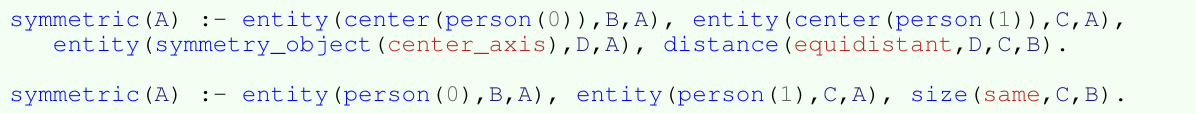}

\smallskip

\myparagraph{Axioms of Perception: Learning Spatio-Temporal Dynamics}
We illustrate learning of spatio-temporal dynamics in the context of visual perception, by learning perceptual patterns from eye-tracking data and people tracks in a movie scene. As an example we focus on attention of a person switching from one individual to another.

\includegraphics[width = \columnwidth]{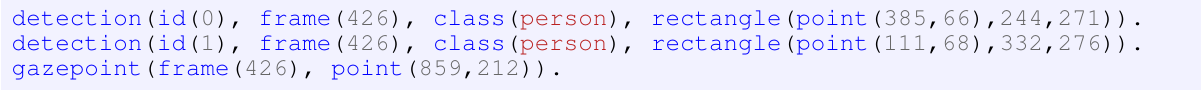}

\smallskip

\emph{Learning: }  \quad We adapt the general learning setup of the example above, for learning spatio-temporal dynamics by introducing the predicate  $\operatorname{\mathsf{holds-in/2}}$ to denote that a spatial relation holds between two entities at a time point.

\includegraphics[width = \columnwidth]{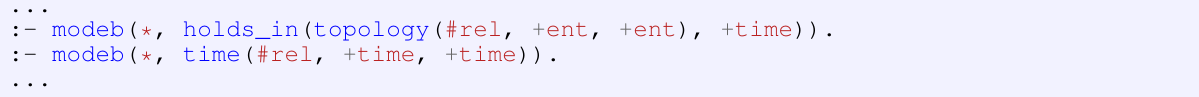}

\smallskip

Spatio-temporal dynamics constituting attention switches include the following.

\includegraphics[width = \columnwidth]{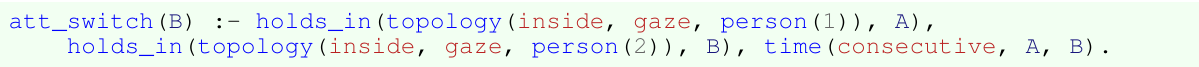}

\mysection{Discussion and Outlook}
Directly comparable to this research is the line of work on integrated inductive-abductive reasoning for learning spatio-temporal relational models from video in \citep{ilp/DubbaBDHC12,jair/DubbaCHBD15}; here, spatio-temporal learning in the context of ILP has only been addressed for the case of topological relations. Furthermore, the ILP learning framework does not have built-in semantics for the topological relations. Aside from this, learning relational spatial structures was investigated in the context of learning spatial relations from language\cite{ilp/Kordjamshidi2011}, and within the geospatial domain \cite{ilp/Malerba2001,ilp/Vaz2010}. Probabilistic Logic Programming frameworks such as PRISM \cite{sato2008new} and ProbLog \cite{kimmig2007probabilistic} have been used for learning parameters, and the structure, of probabilistic logic programs, although (qualitative) spatial reasoning has not been directly addressed. The main point-of-departure of this paper with respect to the state of the art in (qualitative) spatial learning is that the semantics of spatial, temporal, and spatio-temporal relations are directly built within the inductive learning framework of ILP. Pragmatically, what this implies is that it is possible to seamlessly decribe a learning problem using a generic relational spatio-temporal ontology directly as part of a logic programming based learning environment. To the best of our knowledge, such a general spatio-temporal learning framework with built in semantics for mixed qualitative-quantitative spatio-temporal reasoning capabilities has not been available before. Furthermore, the ontology of space-time features supported in our framework goes much beyond topological relations addressing orientation, distance, and size. Future research will focus on enhancing the expressivity of the spatio-temporal relations to cover a wider range of domain-independent features characterising spatio-temporal dynamics.

\small
\nocite{Muggleton94inductivelogic,aleph-ilp-system}
\bibliographystyle{abbrvnat}
%\bibliography{paper-assets/ilp-2016-bib-short}

\begin{thebibliography}{0}
\providecommand{\natexlab}[1]{#1}
\providecommand{\url}[1]{\texttt{#1}}
\expandafter\ifx\csname urlstyle\endcsname\relax
  \providecommand{\doi}[1]{doi: #1}\else
  \providecommand{\doi}{doi: \begingroup \urlstyle{rm}\Url}\fi

\end{thebibliography}


\begin{thebibliography}{27}
\providecommand{\natexlab}[1]{#1}
\providecommand{\url}[1]{\texttt{#1}}
\expandafter\ifx\csname urlstyle\endcsname\relax
  \providecommand{\doi}[1]{doi: #1}\else
  \providecommand{\doi}{doi: \begingroup \urlstyle{rm}\Url}\fi

\bibitem[Bhatt(2012)]{Bhatt:RSAC:2012}
M.~Bhatt.
\newblock {Reasoning about Space, Actions and Change: A Paradigm for
  Applications of Spatial Reasoning}.
\newblock In \emph{Qualitative Spatial Representation and Reasoning: Trends and
  Future Directions}. IGI Global, USA, 2012.
\newblock ISBN ISBN13: 9781616928681.

\bibitem[Bhatt and Loke(2008)]{bhatt:scc:08}
M.~Bhatt and S.~Loke.
\newblock {Modelling Dynamic Spatial Systems in the Situation Calculus}.
\newblock \emph{Spatial Cognition and Computation}, 8\penalty0 (1):\penalty0
  86--130, 2008.
\newblock ISSN 1387-5868.

\bibitem[Bhatt et~al.(2009)Bhatt, Dylla, and Hois]{cosit09/BhattDH09}
M.~Bhatt, F.~Dylla, and J.~Hois.
\newblock Spatio-terminological inference for the design of ambient
  environments.
\newblock In \emph{Spatial Information Theory, 9th International Conference,
  {COSIT} 2009, Aber Wrac'h, France, September 21-25, 2009, Proceedings},
  volume 5756 of \emph{Lecture Notes in Computer Science}, pages 371--391.
  Springer, 2009.
\newblock ISBN 978-3-642-03831-0.
\newblock \doi{10.1007/978-3-642-03832-7_23}.

\bibitem[Bhatt et~al.(2011)Bhatt, Lee, and Schultz]{clpqs-cosit11}
M.~Bhatt, J.~H. Lee, and C.~P.~L. Schultz.
\newblock {CLP(QS):} {A} declarative spatial reasoning framework.
\newblock In \emph{Spatial Information Theory - 10th International Conference,
  {COSIT} 2011, Belfast, ME, USA, September 12-16, 2011. Proceedings}, pages
  210--230, 2011.
\newblock \doi{10.1007/978-3-642-23196-4_12}.

\bibitem[Dubba et~al.(2011)Dubba, Bhatt, Dylla, Hogg, and
  Cohn]{ilp/DubbaBDHC12}
K.~S.~R. Dubba, M.~Bhatt, F.~Dylla, D.~C. Hogg, and A.~G. Cohn.
\newblock Interleaved inductive-abductive reasoning for learning complex event
  models.
\newblock In S.~Muggleton, A.~Tamaddoni{-}Nezhad, and F.~A. Lisi, editors,
  \emph{Inductive Logic Programming - 21st International Conference, {ILP}
  2011, Windsor Great Park, UK, July 31 - August 3, 2011, Revised Selected
  Papers}, volume 7207 of \emph{Lecture Notes in Computer Science}, pages
  113--129. Springer, 2011.
\newblock ISBN 978-3-642-31950-1.
\newblock \doi{10.1007/978-3-642-31951-8_14}.

\bibitem[Dubba et~al.(2015)Dubba, Cohn, Hogg, Bhatt, and
  Dylla]{jair/DubbaCHBD15}
K.~S.~R. Dubba, A.~G. Cohn, D.~C. Hogg, M.~Bhatt, and F.~Dylla.
\newblock Learning relational event models from video.
\newblock \emph{J. Artif. Intell. Res. {(JAIR)}}, 53:\penalty0 41--90, 2015.
\newblock \doi{10.1613/jair.4395}.

\bibitem[Guesgen(1989)]{guesgen1989spatial}
H.~W. Guesgen.
\newblock \emph{{Spatial reasoning based on Allen's temporal logic}}.
\newblock Technical Report TR-89-049. International Computer Science Institute
  Berkeley, 1989.

\bibitem[Hayes(1985)]{Hayes:Naive-I}
P.~J. Hayes.
\newblock {Naive physics {I}: ontology for liquids}.
\newblock In J.~R. Hubbs and R.~C. Moore, editors, \emph{Formal Theories of the
  Commonsense World}. Ablex Publishing Corporation, Norwood, NJ, 1985.

\bibitem[Hazarika(2005)]{hazarika2005qualitative}
S.~M. Hazarika.
\newblock \emph{Qualitative spatial change: space-time histories and
  continuity}.
\newblock PhD thesis, The University of Leeds, 2005.

\bibitem[Hern{\'a}ndez et~al.(1995)Hern{\'a}ndez, Clementini, and
  Di~Felice]{hernandez1995qualitative}
D.~Hern{\'a}ndez, E.~Clementini, and P.~Di~Felice.
\newblock \emph{Qualitative distances}.
\newblock Springer, 1995.

\bibitem[Kimmig et~al.(2007)Kimmig, De~Raedt, and
  Toivonen]{kimmig2007probabilistic}
A.~Kimmig, L.~De~Raedt, and H.~Toivonen.
\newblock Probabilistic explanation based learning.
\newblock In \emph{European Conference on Machine Learning}, pages 176--187.
  Springer, 2007.

\bibitem[Kordjamshidi et~al.(2011)Kordjamshidi, Frasconi, van Otterlo, Moens,
  and Raedt]{ilp/Kordjamshidi2011}
P.~Kordjamshidi, P.~Frasconi, M.~van Otterlo, M.~Moens, and L.~D. Raedt.
\newblock Relational learning for spatial relation extraction from natural
  language.
\newblock In S.~Muggleton, A.~Tamaddoni{-}Nezhad, and F.~A. Lisi, editors,
  \emph{Inductive Logic Programming - 21st International Conference, {ILP}
  2011, Windsor Great Park, UK, July 31 - August 3, 2011, Revised Selected
  Papers}, volume 7207 of \emph{Lecture Notes in Computer Science}, pages
  204--220. Springer, 2011.
\newblock \doi{10.1007/978-3-642-31951-8_20}.

\bibitem[Malerba and Lisi(2001)]{ilp/Malerba2001}
D.~Malerba and F.~A. Lisi.
\newblock Discovering associations between spatial objects: An ilp application.
\newblock In \emph{Proceedings of the 11th International Conference on
  Inductive Logic Programming}, ILP '01, pages 156--163, London, UK, UK, 2001.
  Springer-Verlag.
\newblock ISBN 3-540-42538-1.

\bibitem[Moratz(2006)]{moratz06_opra-ecai}
R.~Moratz.
\newblock Representing relative direction as a binary relation of oriented
  points.
\newblock In \emph{ECAI}, pages 407--411, 2006.

\bibitem[Muggleton and Raedt(1994)]{Muggleton94inductivelogic}
S.~Muggleton and L.~D. Raedt.
\newblock Inductive logic programming: Theory and methods.
\newblock \emph{JOURNAL OF LOGIC PROGRAMMING}, 19\penalty0 (20):\penalty0
  629--679, 1994.

\bibitem[Randell et~al.(1992)Randell, Cui, and Cohn]{randell1992spatial}
D.~A. Randell, Z.~Cui, and A.~G. Cohn.
\newblock A spatial logic based on regions and connection.
\newblock \emph{KR}, 92:\penalty0 165--176, 1992.

\bibitem[Sato and Kameya(2008)]{sato2008new}
T.~Sato and Y.~Kameya.
\newblock New advances in logic-based probabilistic modeling by prism.
\newblock In \emph{Probabilistic inductive logic programming}, pages 118--155.
  Springer, 2008.

\bibitem[Schultz and Bhatt(2016)]{SchultzRuleML2016}
C.~Schultz and M.~Bhatt.
\newblock A numerical optimisation based characterisation of spatial reasoning.
\newblock In \emph{Rule Technologies. Research, Tools, and Applications: 10th
  International Symposium, RuleML 2016, Stony Brook, NY, USA, July 6-9, 2016.
  Proceedings}, pages 199--207. Springer International Publishing, 2016.
\newblock ISBN 978-3-319-42019-6.

\bibitem[Schultz et~al.(2016)Schultz, Bhatt, and Suchan]{sum2016}
C.~Schultz, M.~Bhatt, and J.~Suchan.
\newblock Probabilistic spatial reasoning in constraint logic programming.
\newblock In \emph{Tenth International Conference on Scalable Uncertainty
  Management (SUM 2016) (to appear)}, 2016.

\bibitem[Scivos and Nebel(2004)]{Scivos2004}
A.~Scivos and B.~Nebel.
\newblock {The Finest of its Class: The Natural, Point-Based Ternary Calculus
  LR for Qualitative Spatial Reasoning}.
\newblock In \emph{C. Freksa et al. (2005), Spatial Cognition IV. Reasoning,
  Action, Interaction: International Conference Spatial Cognition. Lecture
  Notes in Computer Science Vol. 3343, Springer, Berlin Heidelberg}, volume
  3343, pages 283--303, 2004.

\bibitem[Srinivasan(2001)]{aleph-ilp-system}
A.~Srinivasan.
\newblock \emph{{The Aleph Manual}}, 2001.
\newblock URL
  \url{http://web.comlab.ox.ac.uk/oucl/research/areas/machlearn/Aleph/}.

\bibitem[Suchan and Bhatt(2016{\natexlab{a}})]{Suchan-2016-IJCAI-Visual}
J.~Suchan and M.~Bhatt.
\newblock Semantic question-answering with video and eye- tracking data -- ai
  foundations for human visual perception driven cognitive film studies.
\newblock In \emph{IJCAI 2016: 25th International Joint Conference on
  Artificial Intelligence}, New York City, USA, July 2016{\natexlab{a}}.

\bibitem[Suchan and Bhatt(2016{\natexlab{b}})]{Suchan-Bhatt-WACV2016}
J.~Suchan and M.~Bhatt.
\newblock The geometry of a scene: On deep semantics for visual perception
  driven cognitive film, studies.
\newblock In \emph{2016 {IEEE} Winter Conference on Applications of Computer
  Vision, {WACV} 2016, Lake Placid, NY, USA, March 7-10, 2016}, pages 1--9.
  {IEEE} Computer Society, 2016{\natexlab{b}}.
\newblock ISBN 978-1-5090-0641-0.
\newblock \doi{10.1109/WACV.2016.7477712}.

\bibitem[Suchan et~al.(2014)Suchan, Bhatt, and Santos]{eccv2014}
J.~Suchan, M.~Bhatt, and P.~E. Santos.
\newblock Perceptual narratives of space and motion for semantic interpretation
  of visual data.
\newblock In \emph{Computer Vision - {ECCV} 2014 Workshops - Zurich,
  Switzerland, September 6-7 and 12, 2014, Proceedings, Part {II}}, pages
  339--354, 2014.
\newblock \doi{10.1007/978-3-319-16181-5_24}.

\bibitem[Suchan et~al.(2016)Suchan, Bhatt, and Yu]{Suchan-ACM-SAP-2016}
J.~Suchan, M.~Bhatt, and S.~Yu.
\newblock The perception of symmetry in the moving image: multi-level
  computational analysis of cinematographic scene structure and its visual
  reception.
\newblock In E.~Jain and S.~J{\"{o}}rg, editors, \emph{Proceedings of the {ACM}
  Symposium on Applied Perception, {SAP} 2016, Anaheim, California, USA, July
  22-23, 2016}, page 142. {ACM}, 2016.
\newblock ISBN 978-1-4503-4383-1.
\newblock \doi{10.1145/2931002.2948721}.

\bibitem[Vaz et~al.(2010)Vaz, Costa, and Ferreira]{ilp/Vaz2010}
D.~Vaz, V.~S. Costa, and M.~Ferreira.
\newblock Fire! firing inductive rules from economic geography for fire risk
  detection.
\newblock In P.~Frasconi and F.~A. Lisi, editors, \emph{Inductive Logic
  Programming - 20th International Conference, {ILP} 2010, Florence, Italy,
  June 27-30, 2010. Revised Papers}, volume 6489 of \emph{Lecture Notes in
  Computer Science}, pages 238--252. Springer, 2010.
\newblock \doi{10.1007/978-3-642-21295-6_27}.

\bibitem[Walega et~al.(2015)Walega, Bhatt, and Schultz]{ampmtqs-lpnmr2015}
P.~A. Walega, M.~Bhatt, and C.~P.~L. Schultz.
\newblock {ASPMT(QS):} {Non-Monotonic Spatial Reasoning with Answer Set
  Programming Modulo Theories}.
\newblock In \emph{Logic Programming and Nonmonotonic Reasoning - 13th
  International Conference, {LPNMR} 2015, Lexington, KY, USA, September 27-30,
  2015. Proceedings}, pages 488--501, 2015.
\newblock \doi{10.1007/978-3-319-23264-5_41}.

\end{thebibliography}

\end{document}